\documentclass{article}

\usepackage{microtype}
\usepackage{graphicx}
\usepackage{subfigure}
\usepackage{booktabs} 

\usepackage{amsmath,amsfonts,bm}

\def\eqref#1{equation~\ref{#1}}

\def\1{\bm{1}}

\DeclareMathAlphabet{\mathsfit}{\encodingdefault}{\sfdefault}{m}{sl}
\SetMathAlphabet{\mathsfit}{bold}{\encodingdefault}{\sfdefault}{bx}{n}

\usepackage{hyperref}

\usepackage[dvipsnames]{xcolor}

\usepackage[accepted]{icml2019}

\icmltitlerunning{Learning Powerful Policies by Using Consistent Dynamics Model}

\begin{document}

\twocolumn[
\icmltitle{Learning Powerful Policies by Using Consistent Dynamics Model}



\icmlsetsymbol{equal}{*}

\begin{icmlauthorlist}
\icmlauthor{Shagun Sodhani}{to}
\icmlauthor{Anirudh Goyal}{to}
\icmlauthor{Tristan Deleu}{to}
\icmlauthor{Yoshua Bengio}{to,so}
\icmlauthor{Sergey Levine}{go}
\icmlauthor{Jian Tang}{to}

\end{icmlauthorlist}

\icmlaffiliation{to}{Mila, University of Montreal}
\icmlaffiliation{so}{CIFAR Senior Fellow}
\icmlaffiliation{go}{University of California, Berkeley}

\icmlcorrespondingauthor{Shagun Shodhani}{sshagunsodhani@gmail.com}

\icmlkeywords{Machine Learning, ICML}

\vskip 0.3in
]



\printAffiliationsAndNotice{} 

\begin{abstract}

Model-based Reinforcement Learning approaches have the promise of being sample efficient. Much of the progress in learning dynamics models in RL has been made by learning models via supervised learning. But traditional model-based approaches lead to ``compounding errors'' when the model is unrolled step by step. Essentially, the state transitions that the learner predicts (by unrolling the model for multiple steps) and the state transitions that the learner experiences (by acting in the environment) may not be \textit{consistent}. There is enough evidence that humans build a model of the environment, not only by observing the environment but also by interacting with the environment. Interaction with the environment allows humans to carry out  \textit{experiments}: taking actions that help uncover true causal relationships which can be used for building better dynamics models. Analogously, we would expect such interactions to be helpful for a learning agent while learning to model the environment dynamics. In this paper, we build upon this intuition by using an auxiliary cost function to ensure consistency between what the agent observes (by acting in the real world) and what it imagines (by acting in the ``learned'' world). We consider several tasks - Mujoco based control tasks and Atari games - and show that the proposed approach helps to train powerful policies and better dynamics models. 
\end{abstract}

\section{Introduction}

Reinforcement Learning consists of two fundamental problems: \emph{learning} and \emph{planning}. \emph{Learning} comprises of improving the agent's current policy by interacting with the environment while \emph{planning} involves improving the policy without interacting with the environment. These problems evolve into the dichotomy of \emph{model-free} methods (which primarily rely on \emph{learning}) and \emph{model-based} methods (which primarily rely on \emph{planning}). Recently, \emph{model-free} methods have shown many successes, such as learning to play Atari games with pixel observations \citep{mnih_dqn_2015, mnih_a3c_2016} and learning complex motion skills from high dimensional inputs \citep{schulman_trpo_2015, schulman_gae_2015}. But their high sample complexity is still a major criticism of the \emph{model-free} approaches. 

In contrast, model-based reinforcement learning methods have been introduced in the literature where the goal is to improve the sample efficiency by learning a dynamics model of the environment.  But model-based RL has several caveats. If the policy takes the learner to an unexplored state in the environment, the learner's model could make errors in estimating the environment dynamics, leading to sub-optimal behavior. This problem is referred to as the model-bias problem \citep{deisenroth2011pilco}.

In order to make a prediction about the future, dynamics models are unrolled step by step which leads to  ``compounding errors'' \citep{talvitie2014model, bengio2015scheduled, lamb_professor_forcing_2016}: an error in modeling the environment at time $t$ affects the predicted observations at all subsequent time-steps. This problem is much more challenging for the environments where the agent observes high-dimensional image inputs (and not compact state representations). On the other hand,  model-free algorithms are not limited by the accuracy of the model, and therefore can achieve better final performance by trial and error, though at the expense of much higher sample complexity. In the model-based approaches, the dynamics model is usually trained with supervised learning techniques and the state transition tuples (collected as the agent acts in the environment) become the supervising dataset. Hence the process of learning the model has no control over what kind of data is produced for its training. That is, from the perspective of learning the dynamics model, the agent just observes the environment and does not  ``interact'' with it. On the other hand, there's enough evidence that humans learn the environment dynamics not just by observing the environment but also by interacting with the environment \citep{cook_where_science_starts_2011, daniels_human_interaction_2015}.  Interaction is useful as it allows the agent to carry out experiments in the real world to determine causality, which is clearly a desirable characteristic when building dynamics models.

\begin{figure*}[h]
\centering
\includegraphics[width=2\columnwidth]{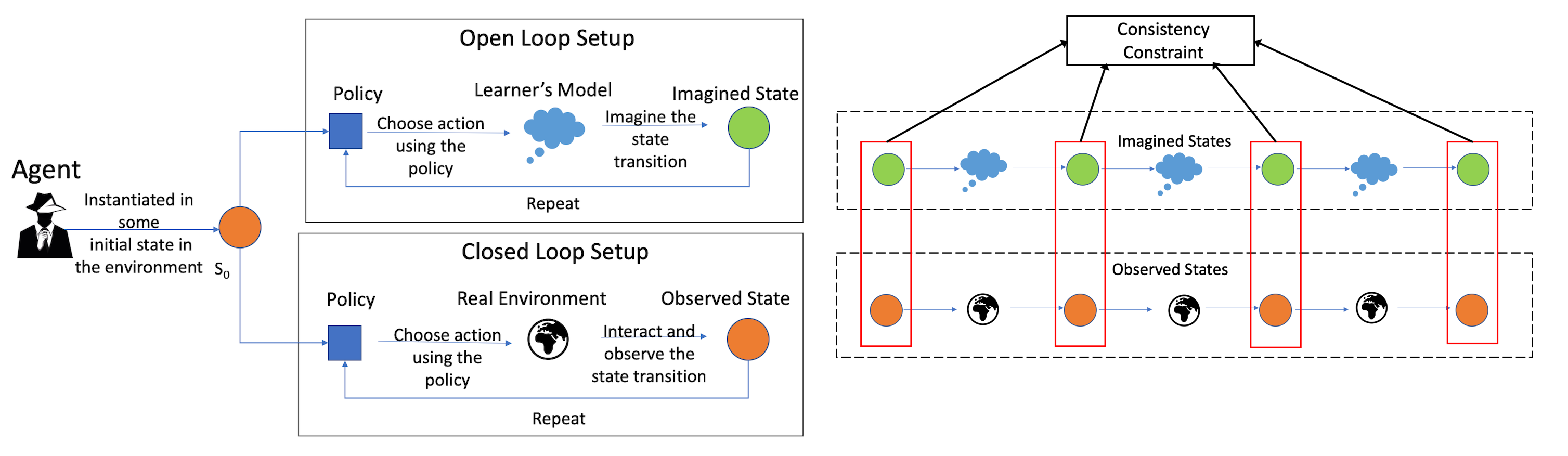}
\caption{The agent, in parallel, (i) Builds a model of the world and (ii) Engages in an interaction with the world. The agent can now learn the model dynamics while interacting with the environment. We show that making these two pathways consistent helps in simultaneously learning a better policy and a more powerful generative model.}
\label{fig:model:loop}
\end{figure*}

This leads to an interesting possibility. The agent could consider two possible pathways: (i) Interacting with the environment by taking actions in the real world to generate new observations and (ii) Interacting with the learned dynamics models by imagining to take actions and predicting the new observations. Consider the humanoid robot from the MuJoCo environment \citep{mordatch_humanoid_2015}. In the first case, the humanoid agent takes an action in the real environment, observes the change in its position (and location), takes another step and so on. In the second case, the agent imagines taking a step, predicts what the observation would look like, imagines taking another step and so on. The first case is the \emph{closed-loop} setup, where the humanoid observes the state of the world, takes an action, gets the true observation from the environment, which it uses to choose the next action, and so on. The second case is the \emph{open-loop} setup, where the agent predicts subsequent states for multiple time steps into the future (with the help of the dynamics model) without interacting with the environment (see figure \ref{fig:model:loop}).

As such, the two pathways may not be \textit{consistent} given the challenges in learning a multi-step dynamics model. By \textit{consistent}, we mean the behavior of state transitions along the two paths should be indistinguishable. Had the predictions from the open loop been similar to the predictions from the closed loop over a long time horizon, the two pathways would be \textit{consistent} and we could say that the learner's dynamics model is grounded in reality. To that end, our contributions are the following:

\begin{enumerate}
    \item We propose to ensure consistency by using an auxiliary loss which explicitly matches the generative behavior (from the open loop) and the observed behavior (from the closed loop) as closely as possible.
    \item We show that the proposed approach helps to simultaneously train more powerful policies as well as better dynamics models, by using a training objective that is not solely focused on predicting the next observation.
    \item We consider various tasks - 7 Mujoco based continuous control tasks and 4 Atari games - from OpenAI Gym suite \citep{1606.01540}, and RLLab \citep{duan_rllab_2016} and show that using the proposed auxiliary loss consistently helps in achieving better performance across all the tasks.
    \item We compare our proposed approach to the state-of-the-art state space models \citep{Buesing_learning_and_querying_2018} and show that the proposed method outperforms the sophisticated baselines despite being very straightforward.
    
\end{enumerate}

We also evaluate our approach on the pixel-based Half-Cheetah task from the OpenAI Gym suite \citep{1606.01540}. The task is difficult for the ``baseline'' state-space models as only the position (and not the velocity) can be inferred from the images, making the task partially observable. Our implementation of the paper is available at \url{https://github.com/shagunsodhani/consistent-dynamics}.
\section{Prelimaries}

A finite time Markov decision process  $\mathcal{M}$ is generally defined by the tuple $(\mathcal{S}, \mathcal{A}, f, R, \gamma)$.
Here, $\mathcal{S}$ is the set of states, $\mathcal{A}$ the action space,   $f(s_{t+1}|s_t, a_t)$ the transition distribution,
$r: \mathcal{S} \times \mathcal{A} \rightarrow R$ is the reward function and $\gamma$ the discount factor. We define the return as the  discounted sum of rewards $r(s_t, a_t)$ along a trajectory $\tau := (s_{0}, a_{0}, ..., s_{T-1}, a_{T-1}, s_{T})$, where $T$ refers to the effective horizon of the process. The goal of reinforcement learning is to find a policy $\pi_\phi$ that maximizes the expected return. Here $\phi$ denotes the parameters of the policy $\pi$.

Model-based RL methods learn the dynamics model from the observed transitions. This is usually done with a function approximator parameterized as a neural network $\hat f_{\theta}(s_{t+1}|s_t, a_t)$. In such a case, the parameters $\theta$ of the dynamics model are optimized to maximize the log-likelihood of the state transition distribution.

\section{Environment Model}

Consider a learning agent training to optimize an expected returns signal in a given environment. At a given timestep $t$, the agent is in some state $s_t \in S$. It takes an action $a_t \in A$ according to its policy  $ a_{t} \sim \pi_t(a_{t}|s_{t})$, receives a reward $r_t$ (from the environment) and transitions to a new state $s_{t+1}$. The agent is trying to maximize its expected returns and has two pathways for improving its behaviour:

\begin{enumerate}
    \item \textbf{\emph{Closed-loop} path:} The learning agent interacts with the environment by taking actions in the real world at every step. The agent starts in state $s_0$ and is in state $s_t$ at time $t$. It chooses an action $a_t$ to perform (using its policy $\pi_t$), performs the chosen action, and receives a reward $r_t$. It then observes the environment to obtain the new state $s_{t+1}$, uses this state to decide which action $a_{t+1}$ to perform next and so on. 
     \item \textbf{\emph{Open-loop} path:} The learning agent interacts with the learned dynamics model by \textit{imagining} to take actions and predicts the future observations (or future belief state in case of state space models). The agent starts in state $s_0$ and is in state $s_t$ at time $t$. Note that the agent ``imagines'' itself to be in state $s_t^I$ and can not access the true state of the environment. It chooses an action $a_t$ to perform (using its policy $\pi_t$), performs the action in the ``learner's'' world (dynamics model) and imagines to transition to the new state $s_{t+1}^I$. Thus the current ``imagined'' state is used to predict the next ``imagined'' state. During these ``imagined'' roll-outs, the agent does not interact with the environment but interacts with its ``learned'' version of the environment which we call the dynamics model or the learner's ``world''.
\end{enumerate}

As an alternative, the agent could use both the pathways simultaneously. The agent could, in parallel, (i) build a model of the environment (\textit{dynamics} model) and (ii) engage in interaction with the real environment as shown in Figure \ref{fig:model:loop}.  We propose to make the two pathways consistent with each other so as to ensure that the predictions from the learner's dynamics model are grounded in the observations from the environment. We show that such a \textit{consistency constraint} helps the agent to learn a powerful policy and a better dynamics model of the environment.  

\subsection{Consistency Constraint}

We want the ``imagined'' behavior (from the open loop) to be consistent with the observed behavior (from the closed loop) to ensure that the predictions from the learner's \textit{dynamics} model are similar to the actual observations from the environment. The \textit{dynamics} model could either be in the observation space (pixel space) or in the state space. State space models are generally more efficient as they model dynamics at a higher level of abstraction. In that case, the learner predicts transitions in the state space by first encoding the actual observation (from the environment) into the state space of the learner and then imposing the consistency constraint in the (learned) state space. 

At a given timestep $t$, the learner is in some environment state $s_t$ while it imagines to be in state $s_t^I$. It takes an action $a_t$ according to its policy  $ a_{t} \sim \pi_t(a_{t}|s_{t})$. Now the learner can make transition in two ways. It could execute the action in the environment and transition to state $s_{t+1}$ (as governed by \textit{f}, the dynamics of the environment). Alternatively, it could execute the action in the learned \textit{dynamics} environment $\hat f_{\theta}$ and imagine to transition to the state $s_{t+1}^I = \hat f_{\theta}(s_{t}^I, a_t)$. Note that the state $s_t$ is not used by the learner's dynamics model when making state transitions during the open-loop setup.

Many possibilities exist for imposing the \textit{consistency constraint}. In this work, we encode the state transitions (during both open-loop and closed-loop) into fixed-size real vectors using recurrent networks and enforce the output of the recurrent networks to be similar in the two cases. Encoding the sequence can be seen as abstracting out the per-step state transitions into how the dynamics of the environment evolve over time. This way, we do not focus on mimicking each state but the high-level dynamics of the state transitions. We encourage the dynamics model to only focus on information that makes the multi-step predictions (from the open-loop) indistinguishable from the actual future observations from the environment (figure \ref{fig:model:loop}). Given the predicted state transitions and the real state transitions, we minimize the $L2$ error between the encoding of predicted future observations as coming from the learner's dynamics model (during open-loop) and the encoding of the future observations as coming from the environment (during closed loop).

Let us assume that the agent started in state $s_0$ and that $a_{0:T-1}$ denote the sequence of actions that the agent takes in the environment from time $t=0$ to $T-1$ resulting in state sequence $s_{1:T}$ that the agent transitions through. Alternatively, the agent could have ``imagined'' a trajectory of state transitions by performing the actions $a_{0:T-1}$ in the learner's dynamics model. This would result in the sequence of states $s^{I}_{1:T}$. The consistency loss is computed as follows:

$$
enc(s_{1:T})) = RNN([s_1, s_2, ..., s_T])
$$
$$
enc(s^I_{1:T})) = RNN([s_1^I, s_2^I, ..., s_T^I])
$$

\begin{equation}
\label{eq::cc_equation}
l_{cc}(\theta, \phi)  = \|enc(s_{1:T})) - enc(s^I_{1:T})) \| 
\end{equation}

where $\|\|$ denotes the L2 norm.

The agent which is trained with the \textit{consistency constraint} is referred to as the \textit{consistent dynamics} agent. The overall loss for such a learning agent can be written as follows:

\begin{equation}
\label{eq::total_loss}
l_{total}(\theta, \phi) = l_{rl}(\phi) + \alpha l_{cc}(\theta, \phi)  
\end{equation}

where $\theta$ refers to the parameters of the agent's transition model $\hat{f}$ and $\phi$ refers to the parameters of the agent's policy $\pi$. The first component of the loss function,  $l_{rl}(\theta, \phi)$, corresponds to the standard RL objective of maximizing the expected return and is referred to as the \textit{RL loss}. The second component of the loss, $l_{cc}(\theta, \phi)$, corresponds to the loss associated with the \textit{consistency constraint} and is referred to as \textit{consistency loss}. $\alpha$ is a hyper-parameter to scale the \textit{consistency loss component} with respect to the \textit{RL loss}.

\subsection{Observation Space Model}
\label{sec::obs::model}
For the observation space models, we represent the environment as a Markov Decision Process $\mathcal{M}$ with an unknown state transition function $f: \mathcal{S} \times \mathcal{A} \rightarrow \mathcal{S}$. At time $t$, the agent is in state $s_t \in \mathcal{S}$, learns a policy function $\pi_{t}$ and a dynamics model $\hat{f}$ to predict the next state $s_{t+1}$ given a state-action pair ($s_t$, $a_t$). We use the hybrid Model-based and Model-free (Mb-Mf) algorithm \citep{nagabandi_mbmf_2017} as the baseline to design and learn the transition function and the policy. \cite{nagabandi_mbmf_2017} propose to use a trained, deep neural network based dynamics model to initialize a model free learning agent to combine the sample efficiency of model-based approaches with the high task-specific performance of model-free methods. Both the transition function and the policy are parameterized using neural networks (Gaussian outputs) as $\hat{f}_{\theta}(s_t, a_t)$ and $\pi_{\phi}(s_t)$ where $\theta$ and $\phi$ denote the parameters of the dynamics model and the policy respectively. The details about model and policy implementation are provided in the appendix.

In the closed loop setup, the agent starts in a state $s_0$. At time $t$, it is in state $s_t$, it chooses an action $a_t \sim \pi_t(a_t | s_t)$, receives a reward $r_t$ and observes the next state $s_{t+1}$ which it uses to choose the next action $a_{t+1}$. In the open loop setup, the agent starts in a state $s_0$. At time $t$, it is in state $s_t$, while it imagines to be in state $s_t^I$. It  chooses an $a_t \sim \pi_t(s_t)$, imagines the next state $s^{I}_{t+1} = \hat f(s_t^I, a_t)$. Simultaneously, the action $a_t$ is simulated in the environment to obtain the next environment state $s_{t+1}$. These environment states are needed to compute the \textit{consistency loss} for training the agent. As described in equation \ref{eq::cc_equation}, we encode the two state transition sequences into fixed size vectors using recurrent models and then minimize the L2 norm between them.

\subsection{State Space Model}
\label{sec::state::model}
If the observation space is high dimensional, as in case of pixel-space observations(from high dimensional image data), state space models may be used to model the dynamics of the environment. These models can be computationally more efficient than the pixel-space models as they make predictions at a higher level of abstraction and learn a compact representation of the observation. Further, it may be easier to model the environment dynamics in the latent space as compared to the high dimensional pixel space. 

We use the state-of-the-art \textit{Learning to Query} model \citep{Buesing_learning_and_querying_2018} as our state space model. Consider a learning agent operating in an environment that produces an observation $o_t$ at every time-step $t$. These observations can be high-dimensional and highly redundant (for modelling the dynamics of the environment). The agent learns to encode these observations $(o_t)$ into compact state-space representations ($s_t$) using an encoder $e$ and learns a policy function $\pi$ to choose actions $a_t\sim\pi(a_t | s_t)$.

The environment dynamics is given by an unknown observation transition function $f: \mathcal{O} \times \mathcal{A}\rightarrow \mathcal{O}$ and the agent aims to learn the model dynamics in state-space representation using a state transition function $\hat{f}$. Both the policy and state transition functions are parameterized using neural networks as $\pi_{\phi}$ and $\hat{f}_{\theta}$, where $\phi$ and $\theta$ represent the parameters of the policy and the transition function respectively.  A latent variable $z_t$ is introduced per timestep to introduce stochasticity in the state transition function. The observation space decoding $o_{t+1}$ can be obtained from the state space encoding as $o_{t+1} \sim p(o_{t+1} | s_t, z_t)$. We now describe the steps in the closed loop and open loop setup. 

\paragraph{Closed Loop:}

The agent starts in some state $s_0$ and receives an observation $o_1$ from the environment. At time $t$, the agent is in a state $s_{t-1}$ and receives an observation $o_t$ from the environment. It samples a latent state vector $z_t \sim q(z_t | e(o_t), s_{t-1}, a_{t-1})$ and transition to a new state, $s_t = {f}_\theta(z_t, s_{t-1}, a_{t-1})$. It selects an action $a_t = \pi(a_t | s_t)$ and decodes the state $s_t$ into observation $o_{t+1} \sim p(o_{t+1} | s_t, z_t)$.

\paragraph{Open Loop:}
The agent starts in some state $s_0$. At time $t$, the agent is in an imagined state $s^I_{t-1}$. It samples a latent state vector $z_t \sim p(z_t | s^I_{t-1}, a_{t-1})$ and transitions to a new imaginary state $s^I_t = \hat{f}_\theta(z_t, s^I_{t-1}, a_{t-1})$. 

When the agent performs the action $a_t$ in the dynamics model, the action is simultaneously simulated in the external environment to obtain the next true observation $o_{t}$. These environment observations are then encoded into the latent state and are needed to ensure consistency between the learner's imagined state transition and the actual state transitions in the real environment. $s^{I}_{1:T}$ denotes the sequence of states that the agent imagines and $o_{1:T}$ denotes the sequence of observations that the agent obtains from the environment. These observations are encoded into the state space to yield a sequence of encoded environment observations $s_{1:T}$.

We want to make the behavior of sequence $s_{1:T}$ indistinguishable from $s^{I}_{1:T}$. We follow the same approach as observation space models where we encode the two state-transition sequences into fixed length vectors using recurrent models and then minimize the L2 norm between them (as described in equation \ref{eq::cc_equation}).  The agent is trained by imitation learning using trajectories sampled using an expert policy. The details about the model and policy implementation are provided in the appendix.

While stochasticity is useful for capturing long term dependencies, most of the latent space models (with stochastic dynamics) are trained with one step ahead predictions and they tend to produce inconsistent predictions when predicting multiple time steps into the future. By using the proposed consistency loss in the latent space, we can enforce that the multi-step predictions be grounded in the observations from the actual environment. Hence, the use of the proposed consistency loss, to improve the long term predictions (as demonstrated empirically), can also be seen as a regularizer.

\section{Rationale Behind Using Consistency Loss}

\begin{figure*}[!htb]
    \includegraphics[width=\linewidth]{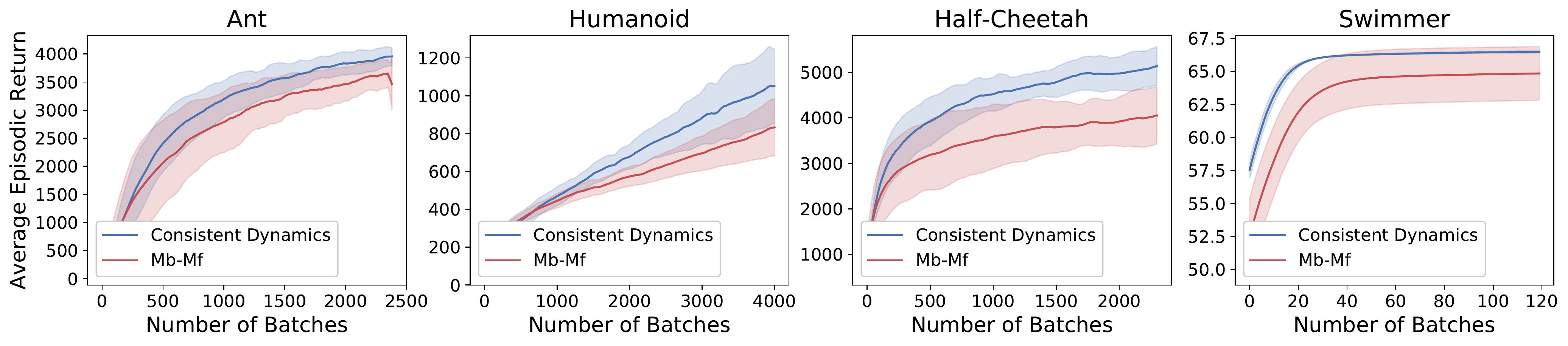}
\caption{Comparison of the average episodic returns, for \textit{Mb-Mf} agent and \textit{consistent dynamics} agent on the Ant, Humanoid, Half-Cheetah and Swimmer environments respectively. Note that the results are averaged over 100 batches for Ant, Humanoid and Half-Cheetah, and 10 batches for Swimmer. }
 \label{fig:obs:average_episodic_return}
\end{figure*}

Our goal is to provide a mechanism for the agent to have a direct ``interaction'' between the agent's policy and its dynamics model. This interaction is different from the standard RL approaches where the trajectories sampled by the policy are used to train the dynamics model. In those cases, the model has no control over what kind of data is produced for its training and there is no (``direct") mechanism for the dynamics model to affect the policy, hence a ``direct interaction'' between the policy and the model is missing.

A practical instantiation of this idea is the \textit{consistency loss} where we ensure consistency between the predictions (from the dynamics model) and the actual observations (from the environment). This simple baseline works surprisingly well compared to the state-of-the-art methods (as demonstrated by our experiments). Applying the consistency constraint means we have two learning signals for the policy: The one from the reinforcement learning loss (to maximize return) and the other due to the consistency constraint.

Our approach is different from just learning a k-step prediction model as in our case, the agent's behavior (i.e the agent's policy) is directly dependent on its dynamics model too. The model and the policy are trained jointly to ensure that the predictions from the dynamics model are consistent with the observation from the environment. This provides a mechanism where learning a model can itself change the policy (thus ``interacting'' with the policy). In the standard case, the policy is optimized only using the RL gradient which aims at maximizing expected reward.  The state transition pairs (collected as the agent acts in the environment) become the supervising dataset for learning the model, and hence the policy is not affected when the model is being updated and there is no feedback from the model learning process to the policy. Hence, the data used for training the model is coming from a policy which is trained independently of how well the model performs on the collected trajectories and the process of learning the model has no control over what kind of data is produced for its training.

\section{Related Work}

\textbf{Model based RL} A large portion of the literature in policy search relies on the  model-free methods, where no prior knowledge of the environment is required to find an optimal policy, through either policy improvement (value-based methods, \cite{rummery1994line,mnih2015human}), or direct policy optimization (policy gradient methods, \cite{mnih_a3c_2016,schulman_trpo_2015}). Although conceptually simple, these algorithms have a high sample complexity. To improve their sample-efficiency, one can learn a model of the environment alongside the policy, to sample experience from.  PILCO \citep{deisenroth2011pilco} is a model-based method that learns a probabilistic model of the dynamics of the environment and incorporates the uncertainty provided by the model for planning on long-term horizons.

This model of the dynamics induces a bias on the policy search though. Previous work has tried to address the model-bias issue of model-based methods, by having a way to characterize the uncertainty of the models, and by learning a more robust policy \citep{deisenroth2011pilco, rajeswaran2016epopt, lim2013reinforcement}. Model Predictive Control \citep[MPC,][]{lenz2015deepmpc} has also been proposed in the literature to account for imperfect models by re-planning at each step, but it suffers from a high computational cost.

There is no sharp separation between model-free and model-based reinforcement learning, and often model-based methods are used in conjunction with model-free algorithms. One of the earliest examples of this interaction is the classic Dyna algorithm \citep{Sutton:1991:DIA:122344.122377}, which takes advantage of the model of the environment to generate simulated experiences, which get included in the training data of a model-free algorithm (like Q-learning, with Dyna-Q). Extensions of Dyna have been proposed \citep{silver2008sample,DBLP:journals/corr/abs-1206-3285}, including deep neural-networks as function approximations. Recently, the Model-assisted Bootstrapped DDPG \citep[MA-DDPG,][]{kalweit2017uncertainty} was proposed to incorporate model-based rollouts into a Deep Deterministic Policy Gradient method. Recently, \citep{weber2017imagination} used a predictive model in Imagination-Augmented Agents to provide additional context to a policy network.

We propose to ensure consistency between the open-loop and the closed-loop pathways as a means to learn a stronger policy, and a better dynamics model. As such, our approach can be applied to a wide range of existing RL setups. Several works have incorporated auxiliary loses which results in representations which can generalize. \cite{unsupervised_aux_tasks} considered pseudo reward functions which help to generalize effectively across different Atari games. In this work, we propose to use the consistency loss for improving both the policy and the dynamics model in the context of reinforcement learning.

\section{Experimental Results}

Our empirical protocol is designed to evaluate how well the proposed \textit{Consistent Dynamics} model compares against the state-of-the-art approaches for observation space models and state space models - in terms of both the sample complexity and the asymptotic performance. We consider Mujoco based environments (observation space models) from RLLab with \cite{nagabandi_mbmf_2017} as the baseline, Mujoco based tasks from OpenAI gym (state space models) with \cite{Buesing_learning_and_querying_2018} as the baseline and Atari games from OpenAI gym with A2C as the baseline. All the results are reported after averaging over 3 random seeds. Note that even though \cite{Buesing_learning_and_querying_2018} is a state-of-the-art model, our simplistic approach outperforms it.

\subsection{Observation Space Models}

We use the hybrid Model-based and Model-free (\textit{Mb-Mf}) algorithm \citep{nagabandi_mbmf_2017} as the baseline model for the observation space models. In this setup, the policy and the dynamics model are learned jointly. The implementation details for these models have been described in the appendix and how to add the consistency loss to the baseline has been described in section \ref{sec::obs::model}. We quantify the advantage of using consistency constraint by considering 4 classical Mujoco environments from RLLab \citep{duan_rllab_2016}: Ant ($S \in R^{41}$, $
A\in R^{8}$), Humanoid ($S \in R^{142}$, $
A\in R^{21}$), Half-Cheetah ($S \in R^{23}$, $
A\in R^{6}$) and Swimmer ($S \in R^{17}$, $
A\in R^{3}$). For computing the consistency loss, the learner's dynamics model is unrolled for $k=20$ steps. The imagined state transitions and the actual state transitions are encoded into fixed length real vectors using GRU \cite{gru}. We report the effect of changing the unrolling length $k$.

\subsubsection{Average Episodic Return}

The average episodic return (and the average discounted episodic return) is a good estimate of the effectiveness of the jointly trained dynamics model and policy. To show that the consistency constraint helps in learning a more powerful policy and a better dynamics model, we compare the average episodic rewards for the baseline \textit{Mb-Mf} model (which does not use the consistency loss) and the proposed \textit{consistent dynamics} model (\textit{Mb-Mf} model + consistency loss). We expect that using consistency would either lead to higher rewards or improve sample efficiency.

Figure~\ref{fig:obs:average_episodic_return} compares the average episodic returns for the agents trained with and without consistency. We observe that using consistency helps to learn a better policy in fewer updates for all the four environments. A similar trend is obtained for the average discounted returns (as shown in the appendix. Since we are learning both the policy and the model of the environment at the same time, these results indicate that using the consistency constraint helps to jointly learn a more powerful policy and a better dynamics model.

\subsubsection{Effect of changing $k$}

During the open-loop setup, the dynamics model is unrolled for $k$ steps. The choice of $k$ could be an important hyper-parameter to control the effect of consistency constraint. 

\begin{figure}[!htb]
    \includegraphics[width=\linewidth]{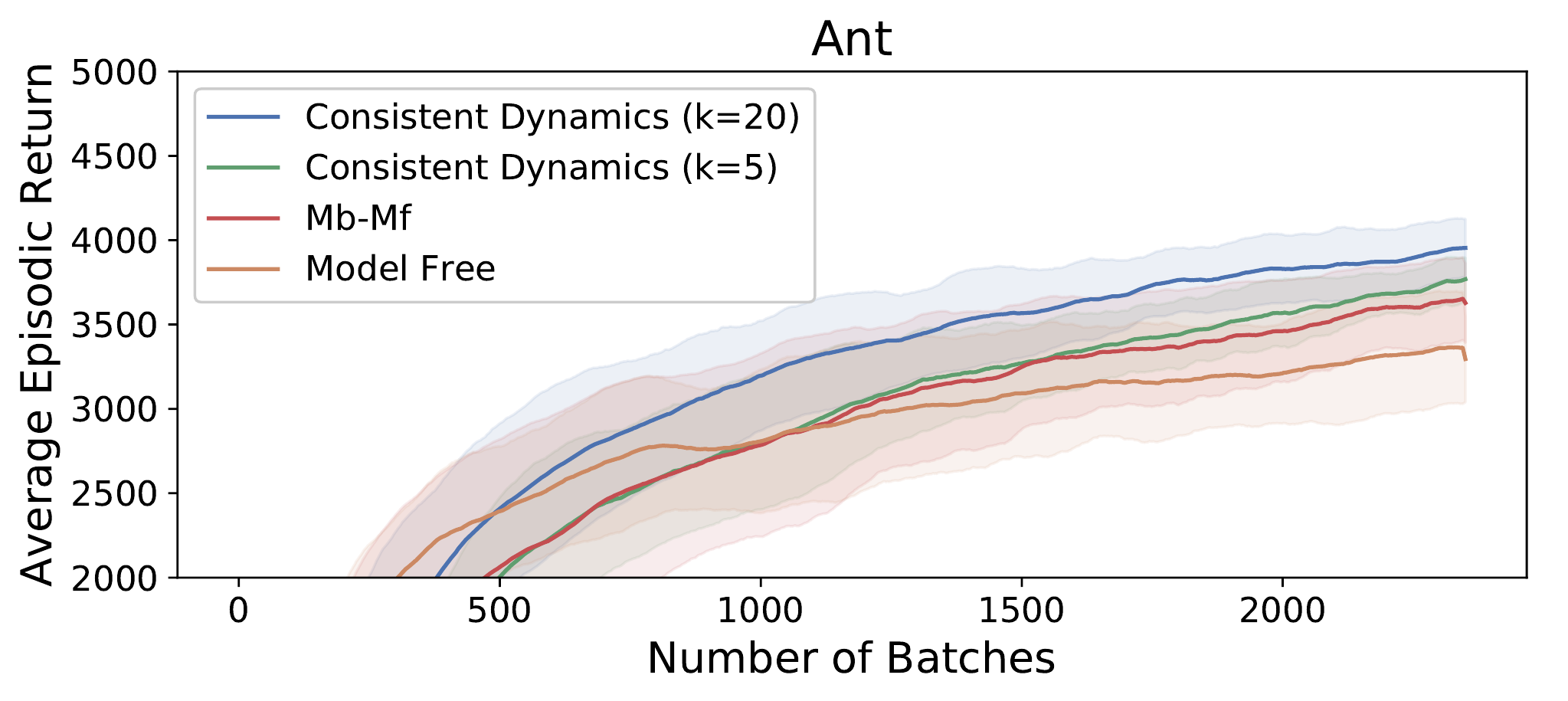}
    \includegraphics[width=\linewidth]{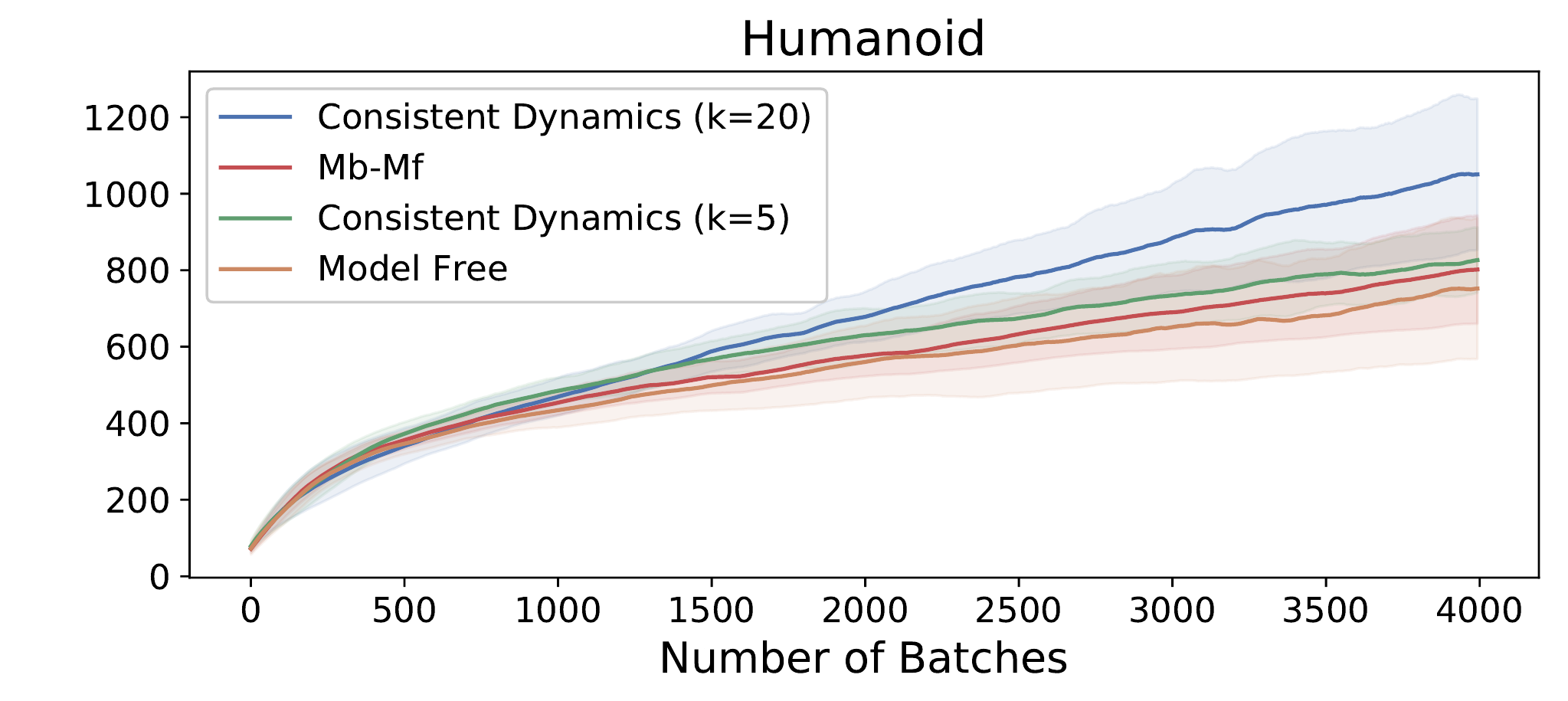}
\caption{Average episodic return on Ant and Humanoid environments, for a model-free agent, the Mb-Mf agent without any consistency constraint, and the Consistent Dynamics (Mb-Mf + consistency constraint) that are trained with a consistency constraint over time horizons of length 5 and 20. Note that the results are averaged over 100 batches for both Ant and Humanoid.}
\label{fig:obs:ablation_study}
\end{figure}

We study the effect of changing $k$ (during training) on the average episodic return for the Ant and Humanoid tasks, by training the agents with $k \in \{5, 20\}$. As an ablation, we also include the case of training the policy without using a model, in a fully model-free fashion. We would expect that a smaller value of $k$ would push the average episodic return of the \textit{consistent dynamics} model closer to the \textit{Mb-Mf} case. Figure~\ref{fig:obs:ablation_study} shows that a  higher value of $k$ ($k=20$) leads to better returns for both tasks.

\subsection{State Space Models}

We use the state-of-the-art \textit{Learning to Query} model \citep{Buesing_learning_and_querying_2018} as the baseline state space model. We train an expert policy for sampling high-reward trajectories from the environment. The trajectories are used to train the policy $\pi_\phi$ using imitation learning and the dynamics model by maximum likelihood. The details about the training setup are described in the appendix and how to add the consistency loss to the baseline has been described in section \ref{sec::state::model}. We consider 3 continuous control tasks from the OpenAI Gym suite \citep{1606.01540}: Half-Cheetah, Fetch-Push \citep{matthias_pusher-robotics_2018} and Reacher. During the open loop, the dynamics model is unrolled for $k=10$ steps for Half-Cheetah and $k=5$ for Fetch-Push and Reacher.

\subsubsection{Evaluating Dynamics Models}

\begin{figure}[!htb]
    \includegraphics[width=\linewidth]{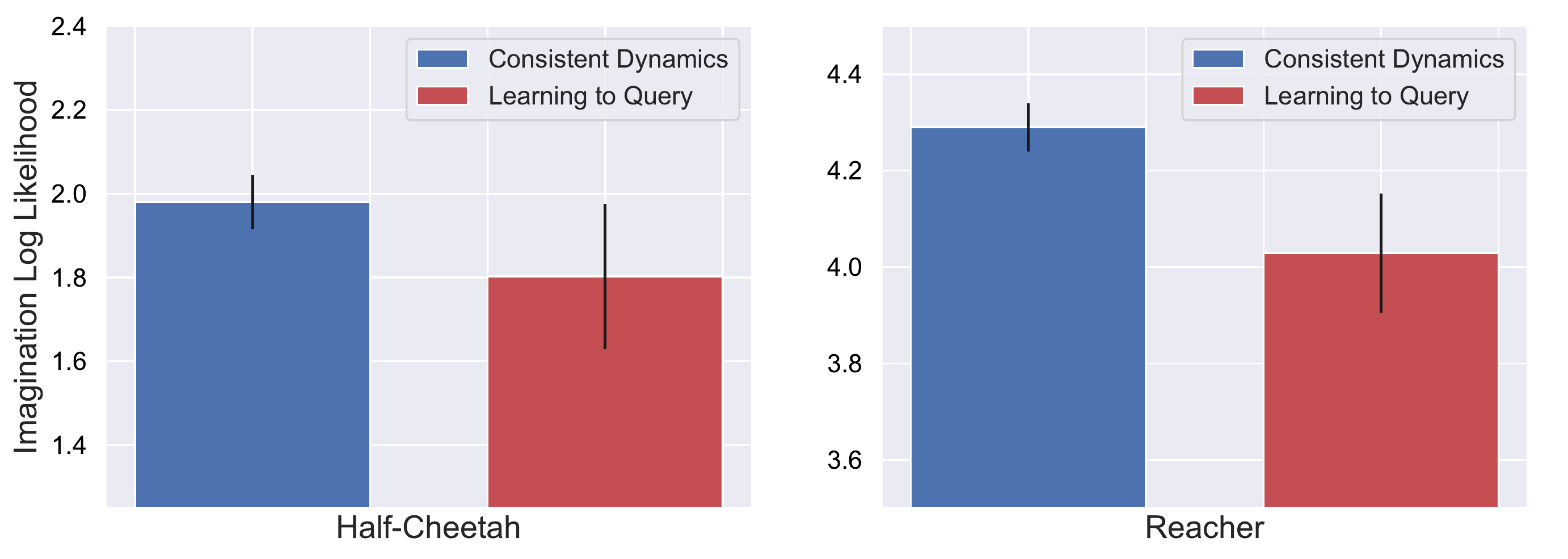}
\caption{Comparison of the imagination log likelihood for the open loop setup for \textit{Consistent Dynamics} agent (ie \textit{Learning to Query} agent + consistency loss) and the baseline (ie \textit{Learning to Query} agent). The plots correspond to
Half-Cheetah and Reacher environments. The plot for Fetch-Push environment is in appendix. The bars represents the values corresponding to the trained agent, averaged over the last 50 batches of training. Consistency constraint leads to a better dynamics model for all the environments.}
\label{fig:state:generative_model:barplot}
\end{figure}

We want to show that the consistency constraint helps to learn a better dynamics model of the environment. Since we learn a dynamics model over the states, we also need to jointly learn an observation model (decoder, see appendix) conditioned on the states. We can then compute the log-likelihood of trajectories in the real environment (sampled with the expert policy) under this observation model. We compare the log-likelihoods corresponding to these observations for the \textit{Learning to Query} agent (trained without the consistency loss) and \textit{Consistent Dynamics} agent (trained with the consistency loss). We expect that the \textit{Consistent Dynamics} agent would achieve a higher log likelihood.

Figure \ref{fig:state:generative_model:barplot} shows that in terms of imagination log likelihood, \textit{Consistent Dynamics} agent (ie \textit{Learning to Query} agent with consistency loss) outperforms the \textit{Learning to Query} agent for all the 3 environments indicating that the agent learns a more powerful dynamics model of the environment. Note that in the case of Fetch-Push and Reacher, we see improvements in the log-likelihood, even though the dynamics model is unrolled for just 5 steps. 

\begin{figure*}[!htb]
\subfigure[Seaquest]
{
  \includegraphics[width=0.5\linewidth]{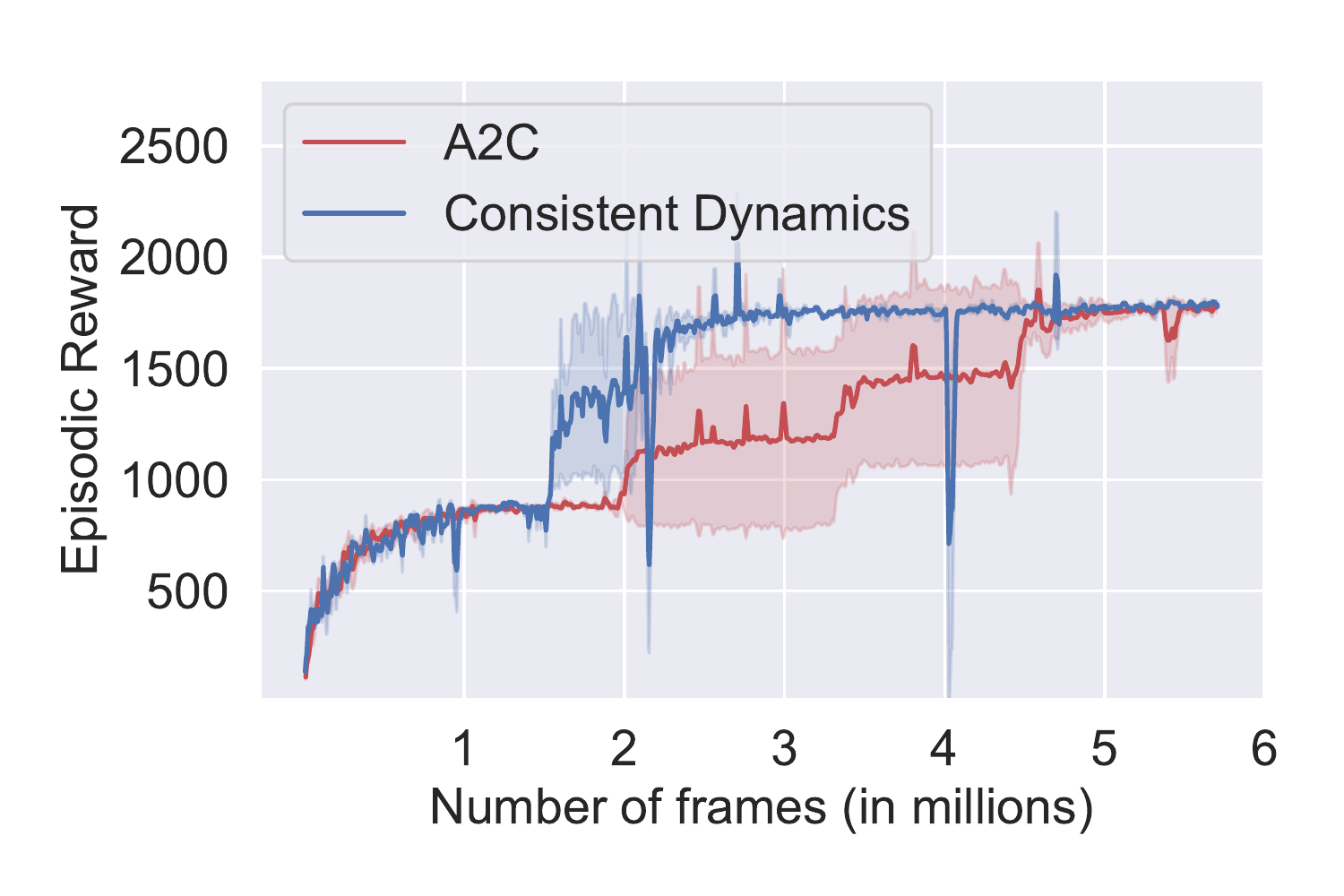}
}
\subfigure[Breakout]
{
  \includegraphics[width=0.5\linewidth]{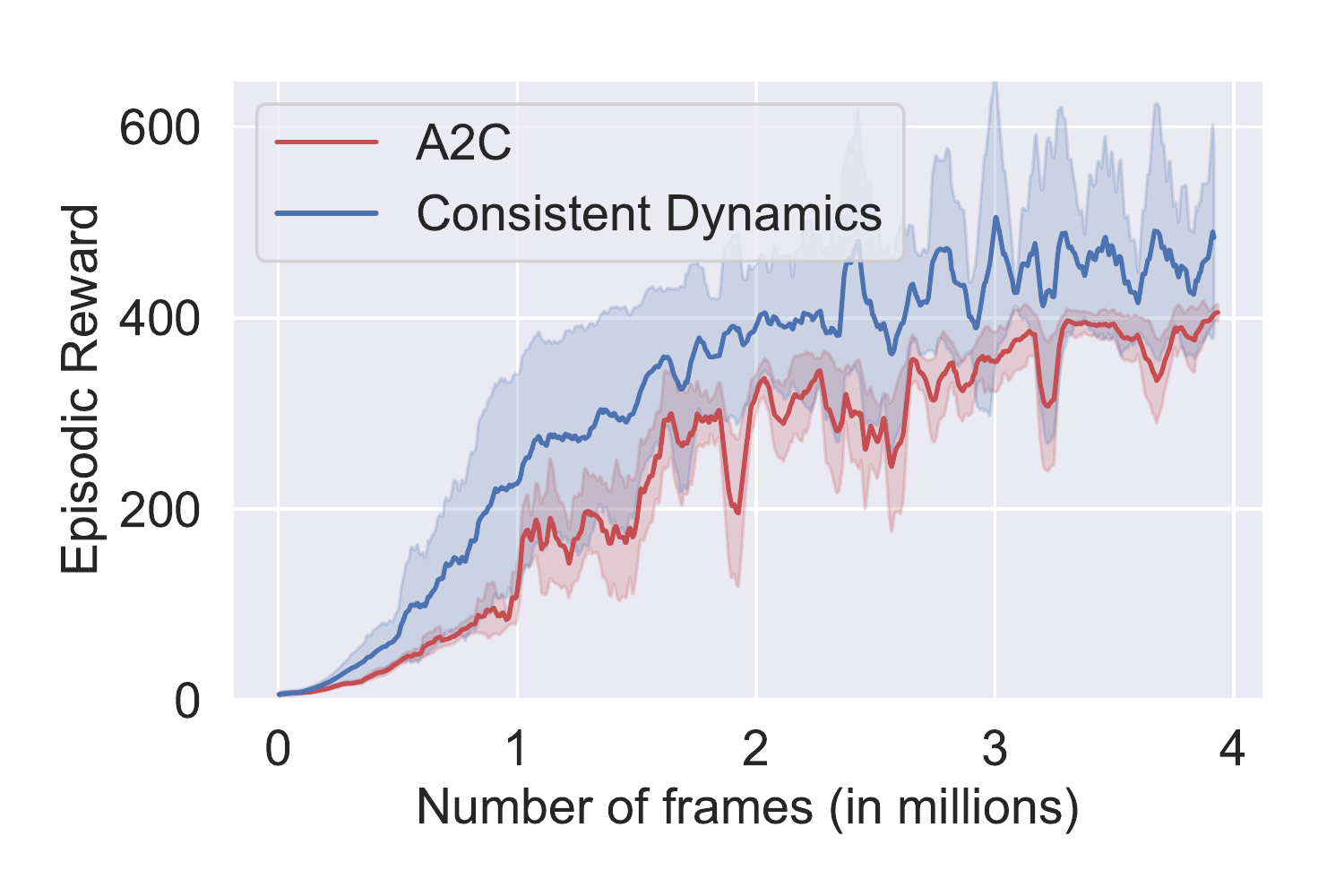}
}
\subfigure[MsPacman]
{
  \includegraphics[width=0.5\linewidth]{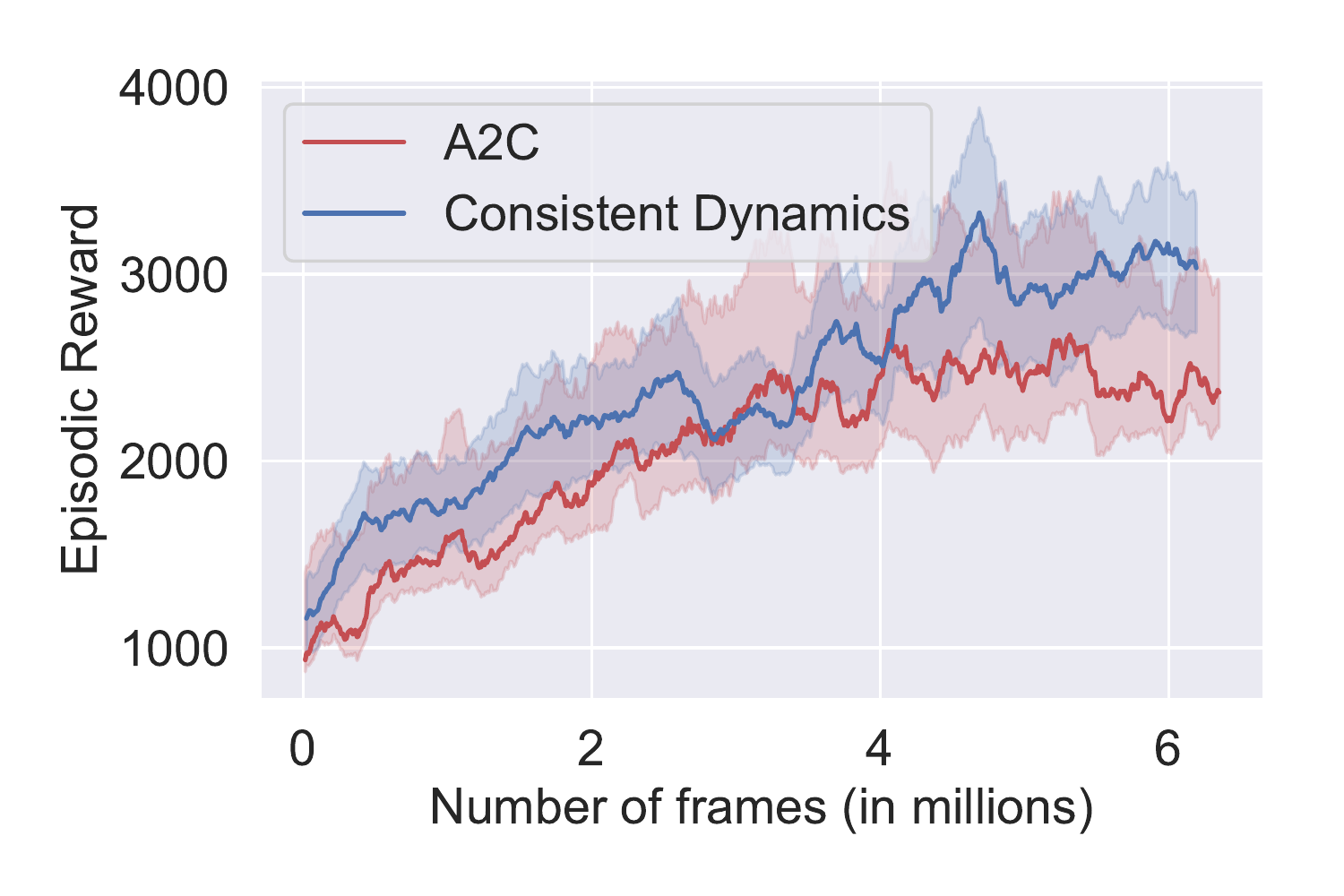}
}
\subfigure[Frostbite]
{
  \includegraphics[width=0.5\linewidth]{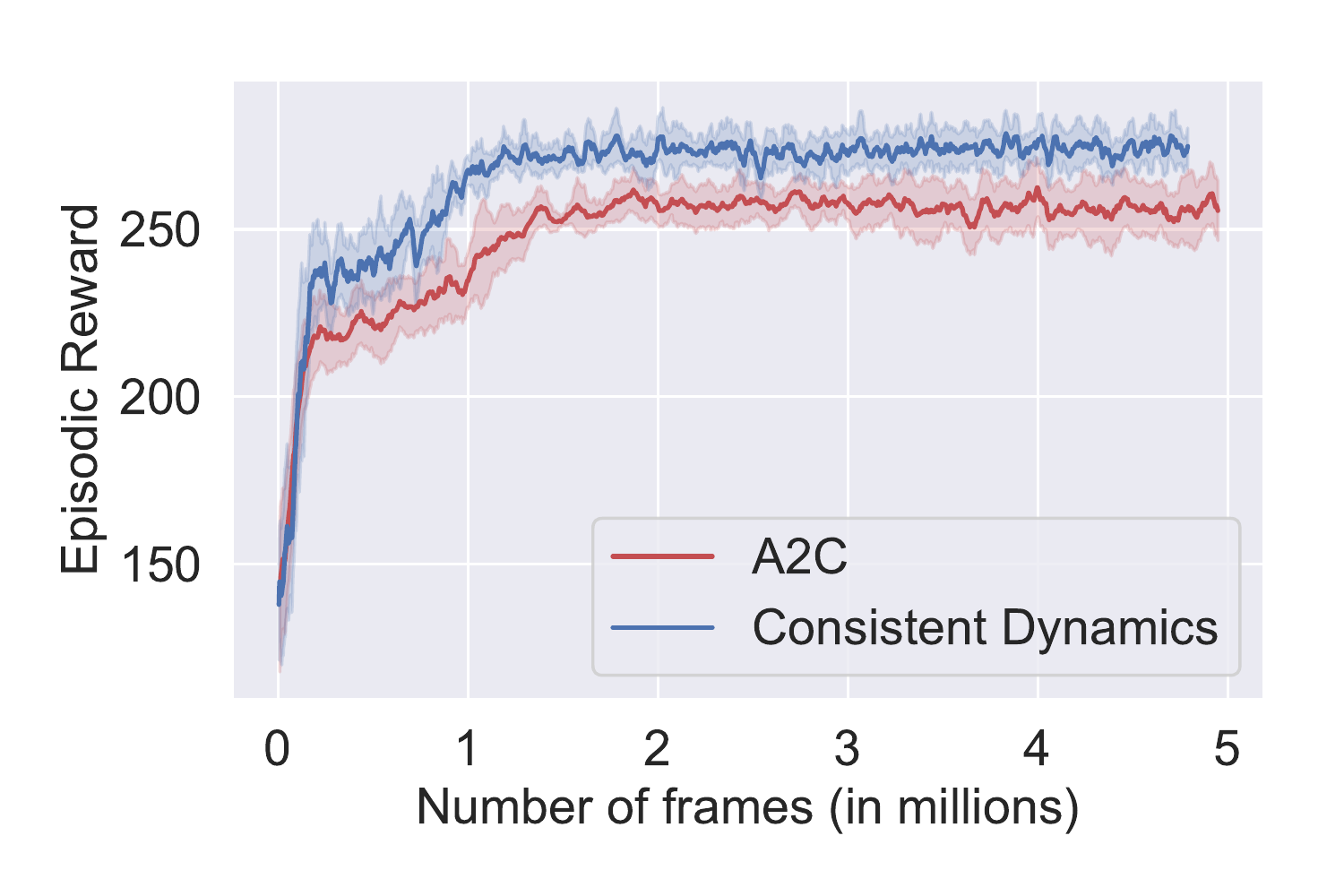}
}
\caption{Comparison of average episodic return on four Atari environments (Seaquest, Breakout, MsPacman and Frostbite respectively), for the \textit{Consistent Dynamics} agent (ie \textit{A2C} agent + consistency loss) and the baseline (just \textit{A2C}). Using consistency constraint leads to a more powerful policy. Note that the results are average over 100 episodes.
}
\label{fig:obs:atari}
\end{figure*}                


\subsubsection{Robustness to Compounding Errors}

\begin{figure}[!htb]
\begin{center}
    \includegraphics[width=\linewidth]{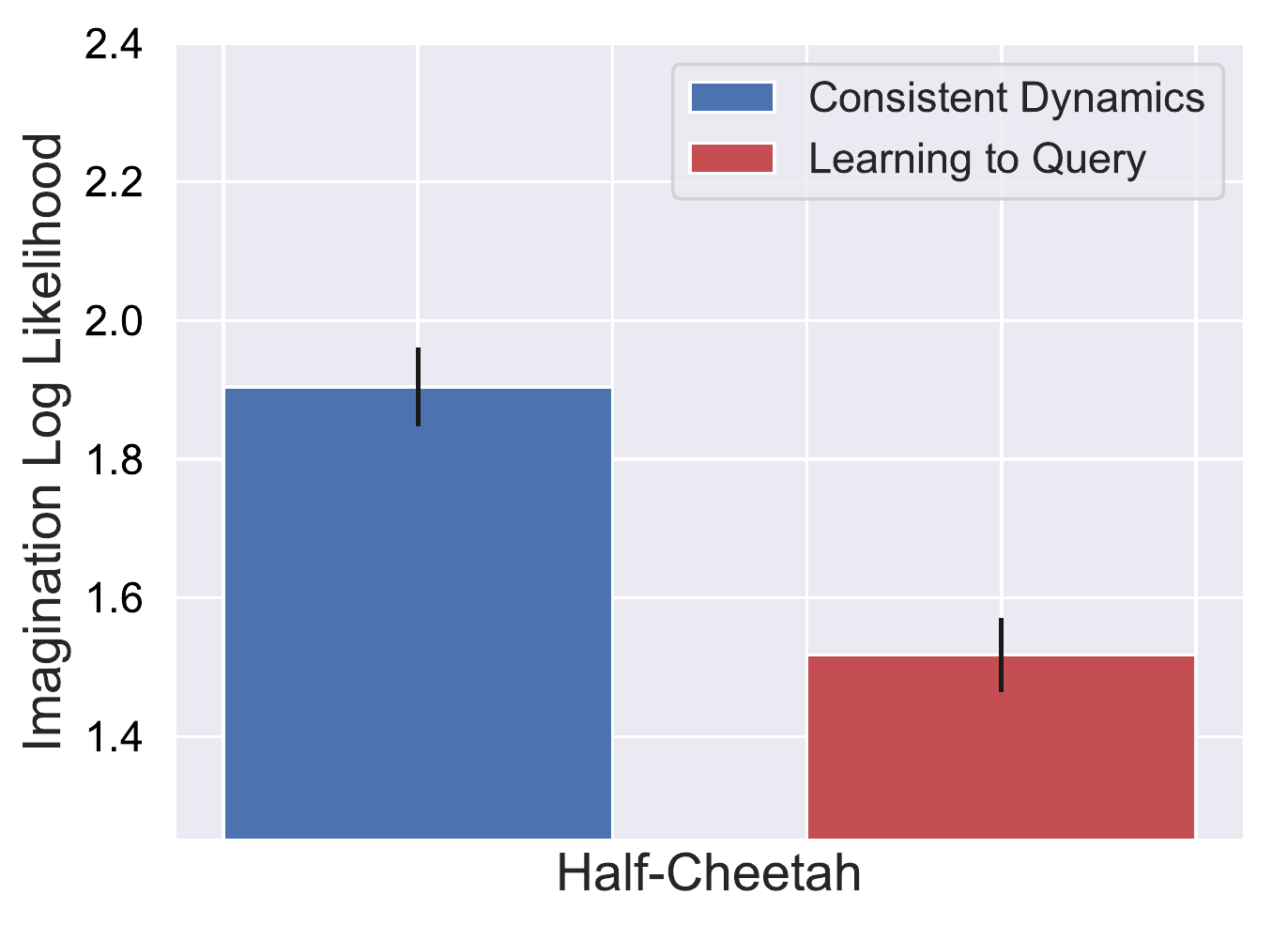}
    \end{center}
\caption{Comparison of the imagination log likelihood for the \textit{Consistent Dynamics} agent and \textit{Learning to Query} agent for Half-Cheetah. The agents were trained with sequence length of $10$ but during testing, the dynamics models were evaluated for length $50$.  The bars represents the values corresponding to the trained agent, averaged over the last 50 batches of training. Using consistency constraint leads to an improved dynamics model (as it achieves better log-likelihood)}
\label{fig:state:generative_model:barplot:ill}
\end{figure}

We also investigate the robustness of the proposed approach in terms of compounding errors. When we use the recurrent dynamics model for prediction, the ground-truth sequence is not available for conditioning. This leads to problems during sampling as even small prediction errors can compound when sampling for a large number of steps.  We evaluate the proposed model for robustness by predicting the future for much longer timesteps (50 timesteps) than it was trained on (10 timesteps). More generally, in figure \ref{fig:state:generative_model:barplot:ill}, we demonstrate that this auxiliary cost helps to learn a better model with improved long-term dependencies by using a training objective that is not solely focused on predicting the next observation, one step at a time.

\subsection{Atari Environment}

We also evaluate our proposed consistency loss on a number of Atari games \cite{bellemare2013arcade} using A2C as the baseline model and by adding consistency loss to A2C to obtain the \textit{Consistent Dynamics} model. Specifically, we consider four environments - Seaquest, Breakout, MsPacman, and Frostbite and
show that in all the 4 environments, the proposed approach is more sample efficient as compared to a vanilla A2C approach thus demonstrating the applicability of our approach to different environments and learning algorithms.

\section{Conclusion}

In this paper, we formulate a way to ensure consistency between the predictions of a dynamics model and the real observations from the environment thus allowing the agent to learn powerful policies, as well as better dynamics models.  The learning agent, in parallel, (i) builds a model of the environment and (ii) engages in an interaction with the environment. This results in two sequences of state transitions: one in the real environment where the agent actually performs actions and other in the agent's dynamics model (or the ``world'') where it imagines taking actions. We apply an auxiliary loss which encourages the behavior of state transitions across the two sequences to be indistinguishable from each other. We evaluate our proposed approach for both observation space models, and state space models and show that the agent learns a more powerful policy and a better generative model. Future work would consider how these two interaction pathways could lead to more targeted exploration. Furthermore, having more flexibility over the length over which we unroll the model could allow the agent to take these decisions over multiple timescales.

\section*{Acknowledgements}

The authors acknowledge the important role played by their colleagues at Mila throughout the duration of this work. The authors would like to thank Bhairav Mehta, Gautham Swaminathan, Koustuv Sinha and Jonathan Binas for their feedback on the initial manuscript. The authors are grateful to NSERC, CIFAR, Google, Samsung, Nuance, IBM, Canada Research Chairs, Canada Graduate Scholarship Program, Nvidia for funding, and Compute Canada for computing resources. We are very grateful to Google for giving Google Cloud credits used in this project.
\bibliography{main}
\bibliographystyle{main}

\newpage
 \section{Appendix}

\subsection{Environment Model}

\subsubsection{Observation Space Model}
\label{app:env:obs}
We use the experimental setup,  environments and the hybrid model-based and model-free (Mb-Mf) algorithm as described in \citep{nagabandi_mbmf_2017}\footnote{Code available here: \href{https://github.com/nagaban2/nn\_dynamics}{https://github.com/nagaban2/nn\_dynamics}}. We consider two training scenarios: training a model-based learning agent with and without the consistency constraint. The consistency constraint is applied by unrolling the model for multiple steps using the observations predicted by the learner's dynamics model (closed-loop setup).  We train an on-policy RL algorithm for Cheetah, Humanoid, Ant and Swimmer tasks from RLLab \citep{duan_rllab_2016} control suite. We report both the average discounted and average un-discounted reward obtained by the learner in the two cases: with and without the use of consistency constraint. The model and policy architectures for the observation space models are as follows:

\begin{enumerate}
    \item \textit{Transition Model}: The transition model $\hat{f}_{\theta}(s_t, a_t)$ has a Gaussian distribution with diagonal covariance, where the mean and covariance are parametrized by MLPs \citep{schulman_trpo_2015}, which maps an observation vector $s_t$ and an action vector $a_t$ to a vector $\mu$ which specifies a distribution over observation space. During training, the log likelihood $p(s|\mu)$ is maximized and state-representations can be sampled from $p(s|\mu)$.
        
    \item \textit{Policy}: The learner's policy $\hat{\pi}_{\phi}(s_t)$ is also a Gaussian MLP which maps an observation vector $s$ to a vector $\mu_{policy}$ which specifies a distribution over action space. Like before, the log-likelihood $p(a|\mu)$ is maximized and actions can be sampled from $p(a|\mu)$.

\end{enumerate}

Learner's policy and the dynamics model are implemented as Gaussian policies with MLPs as function approximations and are trained using TRPO \citep{schulman_trpo_2015}. Following the hybrid Mb-Mf approach \citep{nagabandi_mbmf_2017}, we normalize the states and actions. The dynamics model is trained to predict the change in state $\Delta s_t$ as it can be difficult to learn the state transition function when the states $s_t$ and $s_{t+1}$ are very similar and the action $a_t$ has a small effect on the output. 

\begin{figure*}[ht]
\centering
\includegraphics[width=\textwidth]{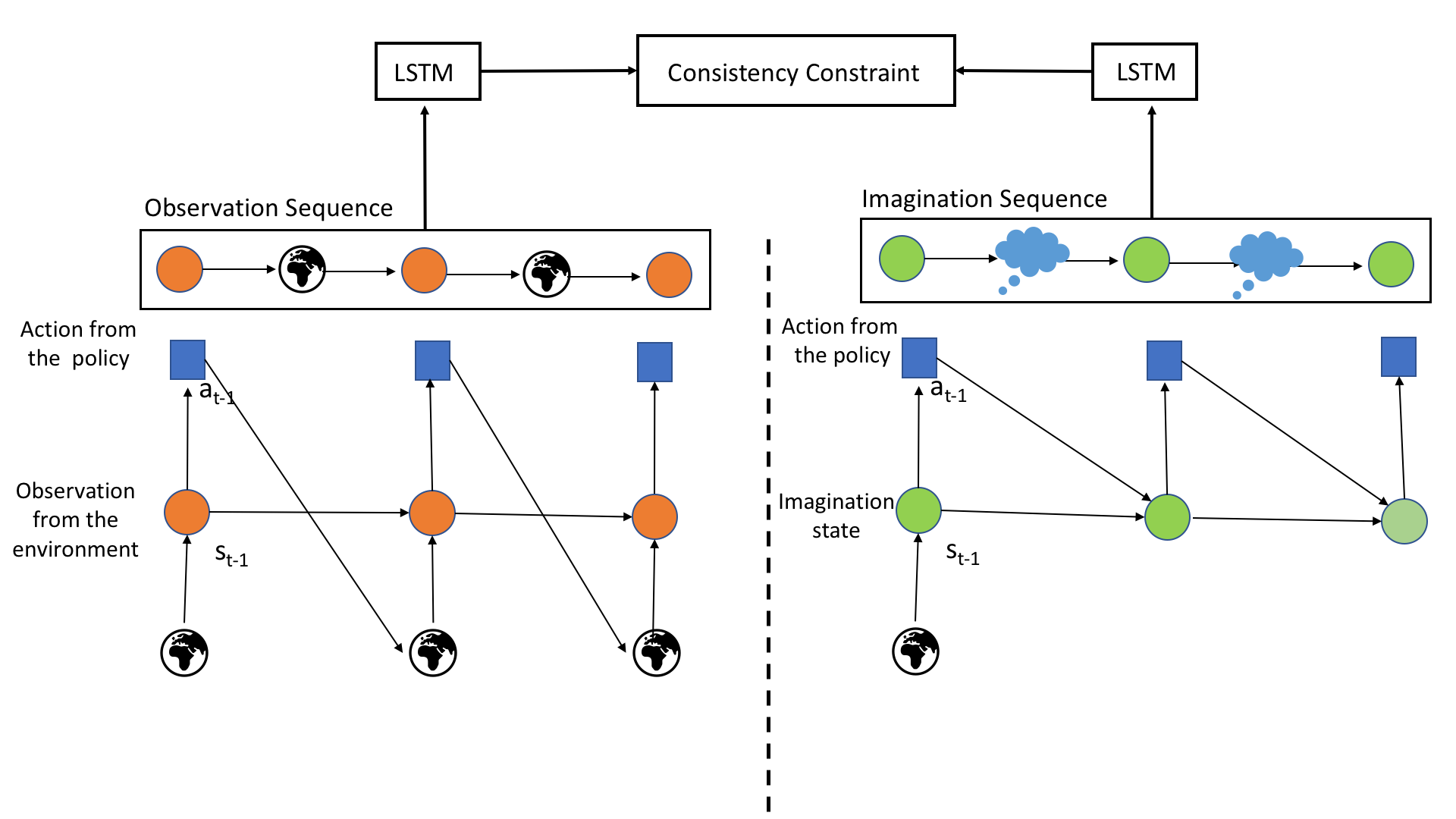}
\caption{Open-loop and closed-loop pathways in the Observation Space Models. The consistency constraint aims to make the behaviour of the open loop predictions indistinguishable from the close loop behaviour}
\label{fig:model:obs}
\end{figure*}

\begin{figure*}[h]
\centering
\includegraphics[width=\textwidth]{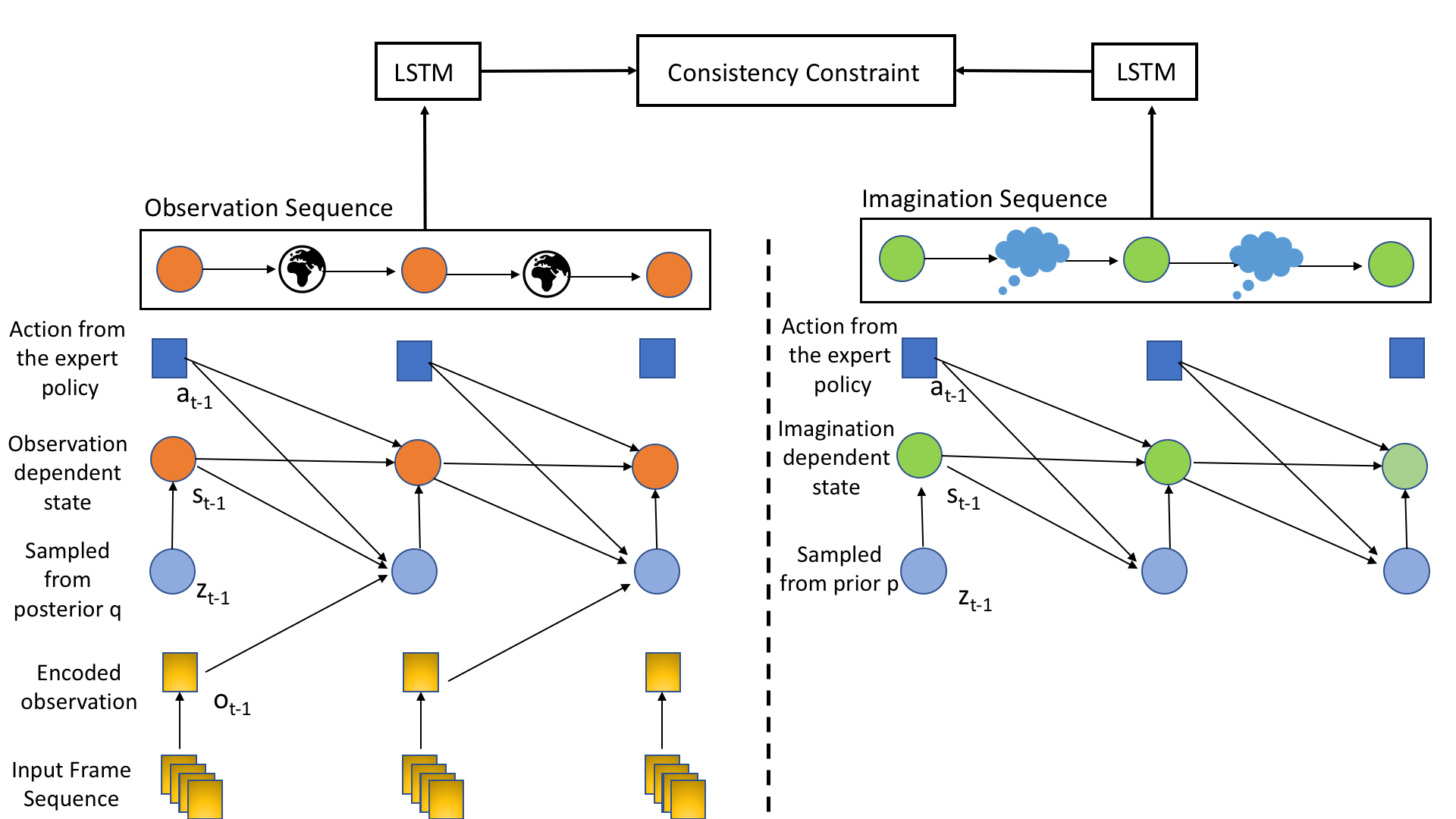}
\caption{Open-loop and closed-loop pathways in the State Space Models. The consistency constraint aims to make the behaviour of the open loop predictions indistinguishable from the close loop behaviour}
\label{fig:model:sst}
\end{figure*}

\subsubsection{State Space Model}
\label{app:env:state-space}
We use the state-of-the-art \textit{Learning to Query} model \citep{Buesing_learning_and_querying_2018} as our state space model.
The model and policy architecture for the state space models are as follows:

\begin{enumerate}
    \item \textit{Encoder}: The learner encodes the pixel-space observations ($64 \times 64 \times 3 $) from the environment into state-space observations ($256$ dimensional vectors) with a convolutional encoder (4 convolutional layers with $4 \times 4$ kernels, stride $2$ and $64$ channels). To model the velocity information, a stack of the latest $4$ frames is used as the observation. The pixel-space observation at time $t-1$ is denoted as $o_{t-1}$, and is encoded into state-space observation $s_{t-1}$.
    \item \textit{Transition Model}: The transition model is a Long Short-Term Memory model \citep[LSTM,][]{hochreiter_lstm_1997}, that predicts the transitions in the state space. For every time-step $t$, latent variables $z_t$ are introduced, whose distribution is a function of previous state-space observation $s_{t-1}$ and previous action $a_{t-1}$. ie $z_t \sim p(z_t | s_{t-1}, a_{t-1})$. The output of the transition model is then a deterministic function of $z_t, s_{t-1},$ and $a_{t-1}$. ie $s_t = f(z_t, s_{t-1}, a_{t-1})$.
    \item \textit{Stochastic Decoder}: The learner can decode the state-space observations back into the pixel-space observations by use of stochastic convolutional decoder. The decoder takes as input the current state-space observation $s_t$ and  the current latent variable $z_t$ and generates the current observation-space distribution from which the learner could sample an observation. ie $o_{t+1} \sim p(o_{t+1} | s_t, z_t)$. This observation model is Gaussian, with a diagonal covariance.
\end{enumerate}

In the closed-loop trajectory, when the learner cannot interact with the environment, the latent variables are sampled from the prior distribution $p(z_t | s_{t-1}, a_{t-1})$. The latent variables are sampled from Normal distributions with diagonal covariance matrices. Since we cannot compute the log-likelihood ${L(\theta)}$ in a closed form for the latent variable models, we minimize the evidence lower bound $\textrm{ELBO}(p_{posterior}) \leq L(\theta)$. As discussed previously, the consistency constraint is applied between the open-loop and closed-loop predictions with the aim of making their behavior as similar as possible. Figure \ref{fig:model:sst} shows a graphical representation of the open-loop and close-loop pathways in the state-space model.

\begin{figure*}[!htb]
    \includegraphics[width=\linewidth]{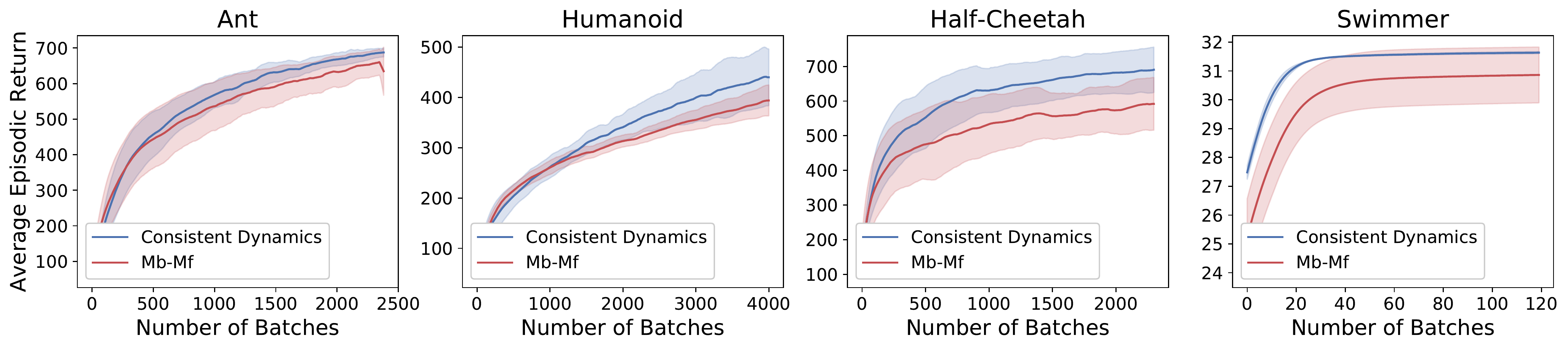}
\caption{Comparison of the average episodic discounted rewards, for agents trained with and without consistency for the Ant, Humanoid, Half-Cheetah and Swimmer environments (respectively). Using consistency constraint leads to better rewards in a fewer number of updates for all the cases. Vertical lines in the rightmost figure show the points of saturation with an equal return. Note that the results are averaged over 100 batches for Ant, Humanoid and Half-Cheetah and 10 batches for Swimmer.}
\label{fig:obs:average_discounted_episodic_return}
\end{figure*}

\paragraph{Expert policy} Having access to some policy trained on a large number of experience is required to sample high-quality trajectories with pixel-observations. To train these expert policies, we used policy-based methods such as Proximal Policy Optimization \citep[PPO,][]{DBLP:journals/corr/SchulmanWDRK17} for the half-cheetah and reacher environments, or Deep Deterministic Policy Gradient with Hindsight Experience Replay \citep[DDPG with HER,][]{NIPS2017_7090} for the pushing task. The architectures and hyper-parameters used are similar to the ones given by the Baselines library \citep{baselines}. Note that these expert policies were trained on the state representation of the agents (ie. the positions and velocities of their joints), while the trajectories were generated with pixel-observations captured from a view external to the agent.

\begin{figure*}[h]
    \includegraphics[width=\linewidth]{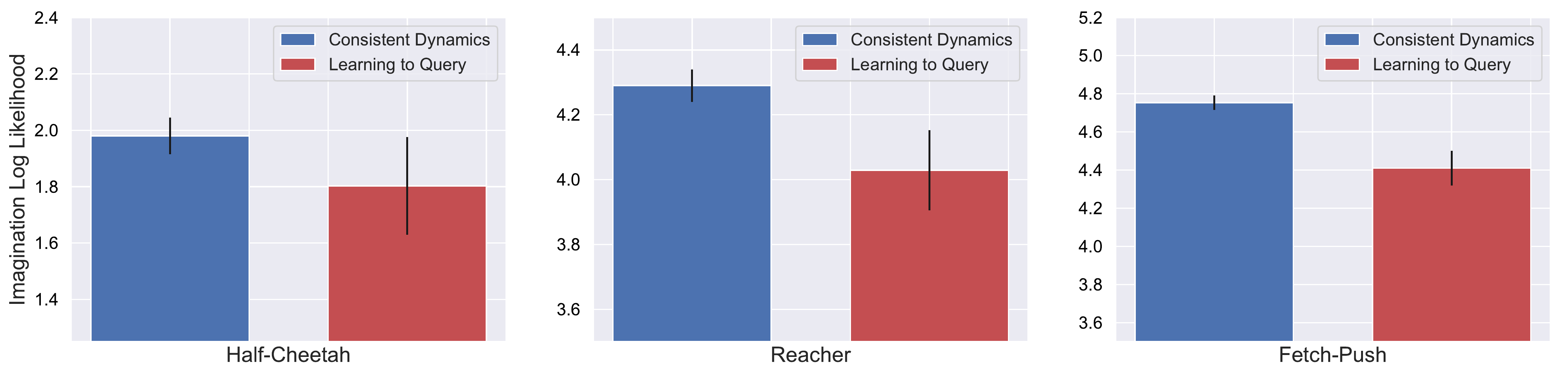}
\caption{Comparison of the imagination log likelihood for the open loop setup for \textit{Consistent Dynamics} agent and \textit{Learning to Query} agent. The plots correspond to
Half-Cheetah, Reacher and Fetch-Push environments respectively. The bars represent the values corresponding to the trained agent, averaged over the last 50 batches of training. Using consistency constraint leads to a better dynamics model for all the 3 environments.}
\label{fig:app:state:generative_model:barplot}
\end{figure*}

\begin{figure*}[h]
    \includegraphics[width=\linewidth]{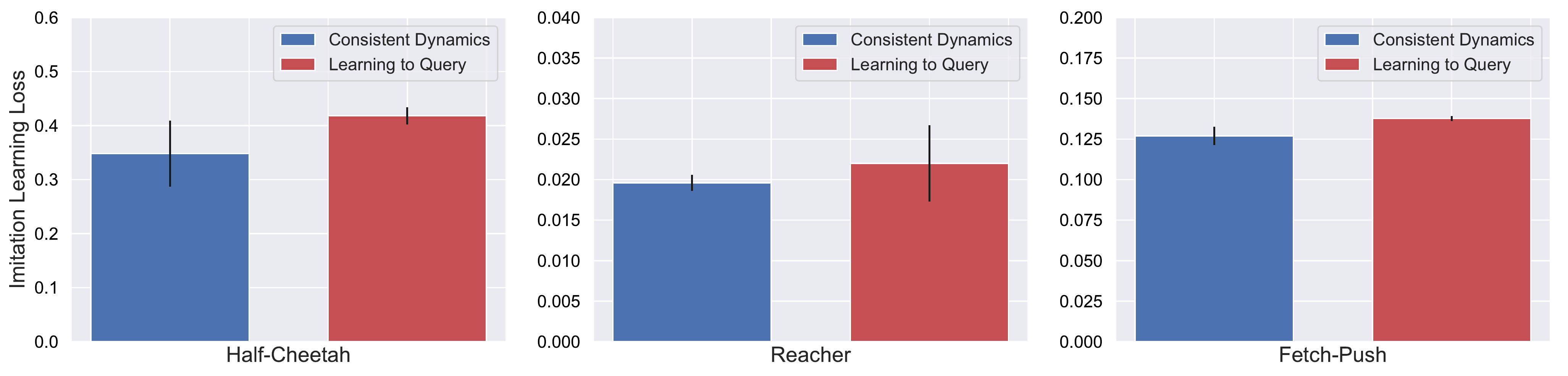}
\caption{Comparison of the imitation learning loss for the \textit{Consistent Dynamics} agent and \textit{Learning to Query} agent. The plots correspond to
Half-Cheetah, Reacher and Fetch-Push environments (respectively). The bars represents the values corresponding to the trained agent, averaged over the last 50 batches of training. Using consistency constraint leads to a more powerful policy.}
\label{fig:app:state:generative_model:barplot2}
\end{figure*}

\subsection{Results}
\subsubsection{Observation Space Models}
\label{app:results:obs}

Figure~\ref{fig:obs:average_discounted_episodic_return} compares the average discounted episodic returns for the agents trained with and without consistency for the observation space models. We observe that using consistency helps to learn a better policy in fewer updates for all the four environments. Since we are learning both the policy and the model of the environment at the same time, these results indicate that using the consistency constraint helps to jointly learn a more powerful policy and a better dynamics model.

\subsubsection{State Space Models}
\label{app:results:obs}

Figure \ref{fig:app:state:generative_model:barplot} shows that in terms of imagination log likelihood, \textit{Consistent Dynamics} agent (ie \textit{Learning to Query} agent with consistency loss) outperforms the \textit{Learning to Query} agent for all the 3 environments indicating that the agent learns a more powerful dynamics model of the environment. Note that in the case of Fetch-Push and Reacher, we see improvements in the log-likelihood, even though the dynamics model is unrolled for just 5 steps. 

For the state-space models, we use the expert trajectories to train our policy  $\pi_\phi$ via imitation learning. To show that consistency constraint helps to learn a more powerful policy, we compare the imitation learning loss for the \textit{Consistent Dynamics} agent (\textit{Learning to Query} agent with consistency loss) and the baseline (\textit{Learning to Query} agent) in figure \ref{fig:app:state:generative_model:barplot2} and observe that the proposed model has a lower imitation learning loss as compared to the baseline model.


\end{document}